\documentclass[letterpaper, 10 pt, conference]{ieeeconf}  

\IEEEoverridecommandlockouts                              

\usepackage{amsmath}
\usepackage{graphicx}
\usepackage[colorinlistoftodos]{todonotes}
\usepackage[colorlinks=true, allcolors=blue]{hyperref}
\usepackage{float}
\usepackage{amssymb}
\usepackage{mathrsfs}  
\usepackage{booktabs}
\usepackage{graphicx}
\usepackage{bbm}
\usepackage{mathtools}

\renewcommand{\phi}{\varphi}
\renewcommand{\epsilon}{\varepsilon}

\usepackage{hybridsystem}

\newcommand{\fmatrix}[1]
{\begin{bmatrix} f{#1}^{r}
(x #1)\\ 
f{#1}^{s}
(x #1) \end{bmatrix}}
\newcommand{\gmatrix}[1]{\begin{bmatrix} g{#1}^{r}_1
(x {#1}) & g{#1}^{r}_2
(x {#1}) \\
    0   
    & g{#1}^{s}
    (x {#1})   \end{bmatrix}}

\newcommand{\dLfy}[1]{\frac{\partial L_{f^{#1} }^{\vec{\gamma}^{#1} - \mathbf{1}} y^{#1}(x)}{\partial x_{#1}} }
\newcommand{\LfLgy}[1]{\frac{\partial L_{f^{#1} }^{\vec{\gamma}^{#1} - \mathbf{1}} y^{#1}(x)}{\partial x_{#1}} g^{#1}(x)}
\newcommand{\Lfy}[1]{\frac{\partial L_{f^{#1}}^{\vec{\gamma}^{#1} - \mathbf{1}} y^{#1}(x)}{\partial x_{#1}} f^{#1}(x)}

\newcommand{\dimSpace}{\nu}  
\newcommand{\dof}{\delta} 
\newcommand{\drs}{\eta} 
\newcommand{\dhl}{\eta} 

\newcommand{\ssc}{\mathrm{ssc}} 

\newcommand{\pt}{pt} 
\newcommand{\pw}{pw} 

\newcommand{\vp}{\vspace{.15cm}}

\title{\LARGE \bf
Control of Separable Subsystems with Application to Prostheses
}

\author{Rachel Gehlhar, Jenna Reher, and Aaron D. Ames
\thanks{This material is based upon work supported by the National Science Foundation Graduate Research Fellowship under Grant No. DGE‐1745301 and NSF NRI Grant No. 1724464.}
\thanks{R. Gehlhar, J. Reher, and A. Ames are with the Department of Mechanical and Civil Engineering, California Institute of Technology, Pasadena, CA 91125 USA. Emails:
{\tt\small $\{$rgehlhar, jreher, ames$\}$@caltech.edu}}}

\begin{document}

\maketitle
\thispagestyle{empty}
\pagestyle{empty}

\begin{abstract}

Nonlinear control methodologies have successfully realized stable human-like walking on powered prostheses. However, these methods are typically restricted to model independent controllers due to the unknown human dynamics acting on the prosthesis. This paper overcomes this restriction by introducing the notion of a \textit{separable subsystem control law}, independent of the full system dynamics. By constructing an equivalent subsystem, we calculate the control law with local information. We build a subsystem model of a general open-chain manipulator to demonstrate the control method's applicability. Employing these methods for an amputee-prosthesis model, we develop a \textit{model dependent} prosthesis controller that relies solely on measurable states and inputs but is equivalent to a controller developed with knowledge of the human dynamics and states.
We demonstrate the results through simulating an amputee-prosthesis system and show the model dependent prosthesis controller performs identically to a feedback linearizing controller based on the whole system, confirming the equivalency.

\end{abstract}


\section{INTRODUCTION}
Powered prostheses offer many advantages over passive prostheses, such as reducing the amputee's metabolic cost \cite{AntagActiveKnee, PowAnkleMetabolicHugh} and increasing the amputee's self-selected walking speed \cite{AntagActiveKnee}. To control these prostheses, previous work focused on dividing a human gait cycle into phases and controlling each with impedance control 
\cite{DesignControlTransProsth},\cite{ConfProsthFive, VirtConsCtrlProst}. These methods are highly heuristic and require extensive tuning. Recent work determined these impedance parameters through a bipedal robot gait generation optimization method \cite{aghasadeghi2013learning}. 
The researchers then improved controller performance by including a feedback term implemented with a model independent quadratic program \cite{zhao2017first}. These successes motivate translating to prostheses other nonlinear control methods proven effective for bipedal robots, such as feedback linearization \cite{ames2014human}, control Lyapunov functions \cite{ames2014rapidly}, and control barrier functions \cite{Nguyen2020}. However two problems arise when trying to translate these methods to prostheses. One, the control laws depend on the full system dynamics, but here the \textit{human dynamics are unknown}. Two, the prosthesis dynamics depend on the full system states, but here the \textit{human states are unknown}.

\begin{figure}[ht] \label{fig:Ampro3} 
\centering
\includegraphics[width=1\columnwidth]{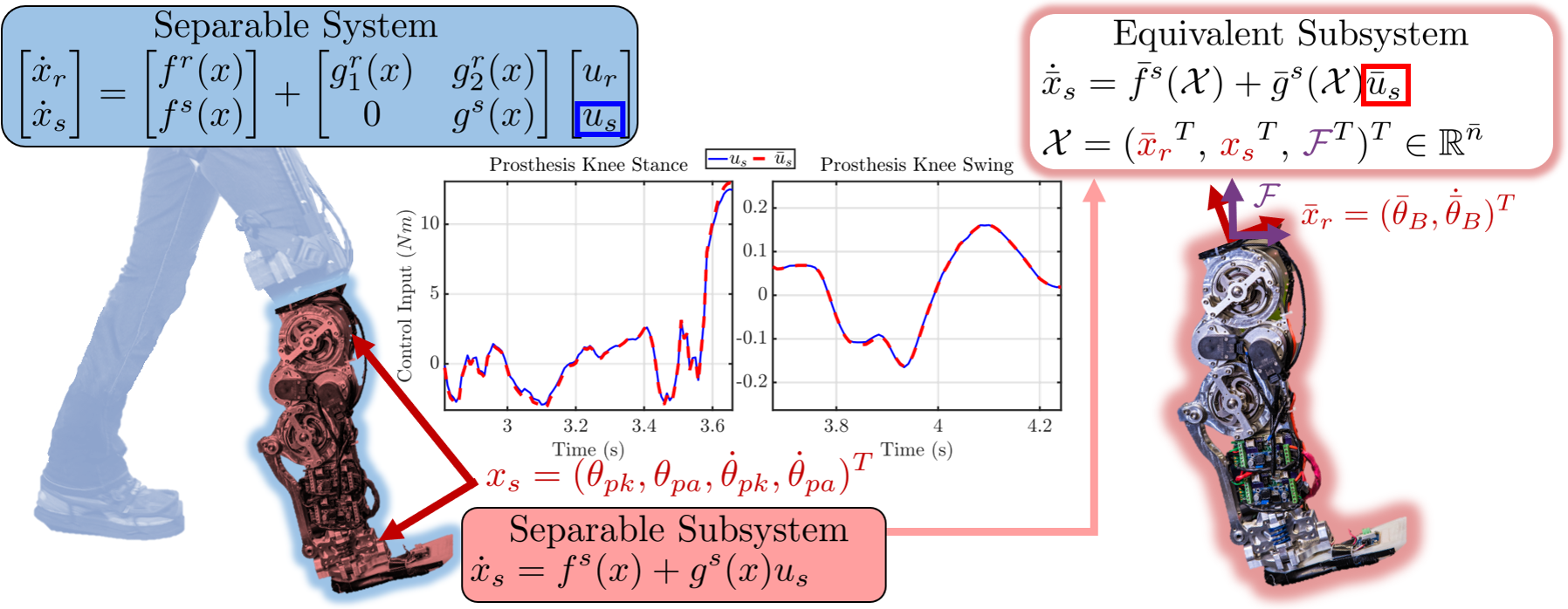}
\vspace{-0.7cm}
{\caption{(Left) Amputee-prosthesis separable system (blue), with separable prosthesis subsystem AMPRO3 (red). (Right) Equivalent prosthesis subsystem. (Middle) Control input from inverse dynamics of human-prosthesis motion capture walking data, determined with full system dynamics (blue) and with equivalent subsystem dynamics (red).}}
\vspace{-0.8cm}
\end{figure}

To examine robotic systems influenced by human behavior, the methods in \cite{ModelContactDynEnvir, DynCtrlSepSyst} considered the interaction forces. This approach is part of a larger investigation of modeling and control of robots in contact with a dynamic environment, which Vukobratovic examined in many of his works, most comprehensively in \cite{DynRobContactTasks}. 
However, these works remain focused on simple models and do not consider incorporating the interaction forces in general nonlinear control methods. The work of \cite{StableRobustHZD} incorporated the interaction forces between the amputee and prosthesis into the prosthesis dynamics to develop a feedback linearizing prosthesis controller. Considering this as a specific example of a subsystem controller, we develop a \textit{general framework to control separable subsystems}.

In Section \ref{sec:SepSubCtrl} we develop a feedback linearizing control law for a \textit{separable subsystem}, a system independent of its full system dynamics. 
While this controller solely depends on the subsystem dynamics, we prove it is \textit{equivalent} to one developed with the full system dynamics, hence guaranteeing full system stability.
Second, we construct the control law using an \textit{equivalent subsystem} with measurable states and inputs. 
In Section \ref{sec:RobCtrl} we examine the application to robotic systems by outlining construction of a subsystem of a general open-chain manipulator.
Following this framework, we model a powered prosthesis, Fig. 1, as a subsystem in Section \ref{sec:AmpProsth}. By using the interaction forces and global orientation and velocities at the amputee's socket, we calculate our \textit{separable subsystem control law}, independent of both the human dynamics and states. We demonstrate the application of these ideas with simulation results in Section \ref{sec:ResultsConclusion}.

These novel methods hold potential to construct model dependent controllers for nonlinear subsystems where the dynamics of the full system are either unknown or computationally expensive. 
This ability could allow nonlinear control approaches to give formal guarantees on stability and safety to coupled subsystems.
This paper approaches subsystem controller construction in the context of both general nonlinear systems and robotic systems to develop proofs of equivalency to the full system feedback linearizing controller, thereby guaranteeing full system stability and providing a general framework applicable to other nonlinear and robotic systems. We develop a prosthesis controller similar to \cite{StableRobustHZD} as a specific example.

\section{SEPARABLE SUBSYSTEM CONTROL} \label{sec:SepSubCtrl}
To analyze controller construction for the amputee-prosthesis system, consider a general affine control system,
\begin{equation} \label{eq:affine}
    \dot{x} = f(x) + g(x) u,
\end{equation}
with states $x \in \mathbb{R}^n$ and control inputs $u \in \mathbb{R}^m$, where $m 
\leq n$. 
The vector fields $f$ and $g$ are locally Lipschitz continuous, meaning given an initial condition $x_0 = x(t_0)$, there exists a unique solution $x(t)$ for some time. For simplicity we assume forward completeness, i.e. solutions exist for all time. 
We can stabilize this system by constructing a \textit{feedback linearizing} control law for $u$, which cancels out the nonlinear dynamics of the system and applies a linear controller to stabilize the resultant linear system.

\newsec{Feedback Linearization.}
To begin construction of $u$ for the affine control system \eqref{eq:affine}, we define linearly independent outputs $y: \mathbb{R}^n \rightarrow \mathbb{R}^m$ so that the system has the same number of outputs as inputs.
These outputs are of vector relative degree  $\vec{\gamma} = (\gamma_1, \gamma_2, \dots, \gamma_m)$, chosen so $\sum_{i = 1}^{m} \gamma_i = n$ such that the system is \textit{full state feedback linearizable} \cite{sastry2013nonlinear}. We define the vector of partial derivatives of the $\gamma_i - 1$ \textit{Lie derivatives} \cite{sastry2013nonlinear, khalil2002nonlinear} of the $y_i(x)$ outputs with respect to the dynamic drift vector $f(x)$ for $i = 1, \dots, m$ as follows:
\begin{equation*}
    \frac{\partial L_f^{\vec{\gamma} - \mathbf{1}} y(x)}{\partial x}
    \triangleq 
    \begin{bmatrix}
    \frac{\partial L_f^{\gamma_1 - 1} y_1(x)}{\partial x} \\
    \frac{\partial L_f^{\gamma_2 - 1} y_2(x)}{\partial x} \\
    \vdots \\
    \frac{\partial L_f^{\gamma_m - 1} y_m(x)}{\partial x}
    \end{bmatrix}.
\end{equation*}
Then, we define the vector of $\gamma_i$ Lie derivatives with respect to drift vector $f(x)$ and control matrix $g(x)$, respectively:
\begin{align*}
    &L_f^{\vec{\gamma}}y(x) 
     \triangleq
    \frac{\partial L_f^{\vec{\gamma} - \mathbf{1}} y(x)}{\partial x} f(x), 
    \\
    &L_gL_f^{\vec{\gamma} - \mathbf{1}} y(x) 
     \triangleq
    \frac{\partial L_f^{\vec{\gamma} - \mathbf{1}} y(x)}{\partial x} g(x). 
\end{align*}
Finally, we construct a feedback linearizing controller,
\begin{equation} \label{eq:FeedLin}
    \begin{aligned}
    u(x) &= -(\underbrace{L_g L_f^{\vec{\gamma} - \mathbf{1}} y(x)}_{A(x)})^{-1}
    (\underbrace{L_f^{\vec{\gamma}} y(x)}_{L_f^*y(x)} - \mu)
    \\
    &= -A^{-1}(x) (L_f^*y(x) - \mu),
    \end{aligned}
\end{equation}

\noindent where $\mu \in \mathbb{R}^m$ is the auxiliary control input the user defines to render the linearized system stable. See \cite{sastry2013nonlinear} or \cite{khalil2002nonlinear} for details. Note that $A(x)$ is invertible because the outputs are linearly independent and the system is square \cite{sastry2013nonlinear}, since the number of inputs equals the number of outputs.

\newsec{Remark 1.}
Here, constructing a feedback linearizing controller requires the dynamics of the full system. However, in the case of large dimensional systems, the full dynamics may be unknown or may become computationally expensive, inhibiting feedback linearization. 

\subsection{Control Law for Separable Subsystem} \label{ssec:CtrlSep}
This section eliminates the need to know the full system dynamics for feedback linearization by constructing a \textit{separable subsystem control law} that only depends on subsystem dynamics.
We begin by defining a \textit{separable control system}.

\vp

\textit{\textbf{Definition 1: }
The affine control system \eqref{eq:affine} is a \textbf{separable control system} if it can be structured as}
\begin{equation} \label{eq:NLform}
\begin{aligned} 
    &\begin{bmatrix}
        \dot{x}_r \\ \dot{x}_s
    \end{bmatrix}
    =
    \fmatrix{} + \gmatrix{}
    \begin{bmatrix}
        u_r \\ u_s
    \end{bmatrix}
    \\
        & x_r \in \mathbb{R}^{n_r}, \quad
        x_s \in \mathbb{R}^{n_s}, \quad
        u_r \in \mathbb{R}^{m_r}, \quad
        u_s \in \mathbb{R}^{m_s},
\end{aligned}
\end{equation}
\textit{where $n_r + n_s = n$ and  $m_r + m_s = m$.}

\vp

\noindent Because of the structure of $g(x)$ in \eqref{eq:NLform}, $u_r$ only acts on part of the system. This motivates defining a \textit{separable subsystem} independent of $u_r$.

\vp

\textit{\textbf{Definition 2. }
For a separable control system \eqref{eq:NLform}, its \textbf{separable subsystem} is defined as}
\begin{equation} \label{eq:Subsystem}
    \dot{x}_s = f^s(x) + g^s(x) u_s,
\end{equation}
\textit{which depends on the full system states $x \in \mathbb{R}^n$.}

\vp

Now, to construct a feedback linearizing control law for this separable subsystem, we construct output functions that solely depend on the subsystem states $x_s \in \mathbb{R}^{n_s}$ and whose Lie derivatives solely depend on the subsystem \eqref{eq:Subsystem}.

\vp

\textit{\textbf{Definition 3. }
For a separable subsystem \eqref{eq:Subsystem} of the separable control system \eqref{eq:NLform}, a set of linearly independent output functions with vector relative degree $\vec{\gamma}^s = (\gamma^s_1, \gamma^s_2, \dots, \gamma^s_{m_s})$ with respect to \eqref{eq:NLform}, where $\sum_{i = 1}^{m_s} \gamma^s_i = n_s$ are \textbf{separable subsystem outputs} if they only depend on $x_s \in \mathbb{R}^{n_s}$,}
\begin{equation} \label{eq:outputs_S}
    y^s(x_s) \in \mathbb{R}^{m_s},
\end{equation}
\textit{and meet the following cross-term cancellation conditions for $j = 1, \dots, \gamma^s_i\ - 1$ and $i = 1, \dots, m_s$:}
\begin{align} \tag{D3.1} \label{conditionSSO1}
    & \frac{\partial L_{f^s}^{j} y^s(x)}{\partial x_r} f^r(x) = 0, 
    \\ \tag{D3.2} \label{conditionSSO2}
    & \frac{\partial L_{f^s}^{\gamma^s_i - 1} y^s(x)}{\partial x_r} 
    \begin{bmatrix}
    g^r_1(x) & g^r_2(x)
    \end{bmatrix}
    =
    \begin{bmatrix}
    0 & 0
    \end{bmatrix}.
\end{align}

\vp

We use these outputs to introduce a \textit{separable subsystem control law} in terms of the subsystem \eqref{eq:Subsystem} alone.

\vp

\textit{\textbf{Definition 4. } For a separable subsystem \eqref{eq:Subsystem} with separable subsystem outputs \eqref{eq:outputs_S}, we define a \textbf{separable subsystem control law} as the feedback linearizing control law}
\begin{equation} \label{eq:us}
    \begin{aligned}
    u_{\ssc}(x) &\triangleq
    - (\underbrace{L_{g^s}L_{f^s}^{\vec{\gamma}^s - \mathbf{1}} y^s(x)}_{A_s(x)})^{-1}
    (\underbrace{L_{f^s}^{\vec{\gamma}^s} y^s(x)}_{L_{f^s}^*y^s(x)}
    - \mu_s)
    \\
    &= -A_s^{-1}(x) (L_{f^{s}}^*y^s(x) - \mu_s).\\
    \end{aligned} 
\end{equation}

\vp

This control law is independent of the rest of the system dynamics $f^r(x),\, g^r_1(x),\, \text{and } g^r_2(x)$, but still depends on the full system states $x$. We will address this dependence in subsequent results to develop an implementable form of this control law solely dependent on \textit{measurable states}.

To compare this control law $u_{\ssc}(x)$ to $u_s(x)$,
we construct \textit{separable outputs} for the full system that include the separable subsystem outputs $y^s(x_s)$ used for \eqref{eq:us}.

\vp

\textit{\textbf{Definition 5. }
For a separable control system, a set of linearly independent output functions with vector relative degree $\vec{\gamma}$ are \textbf{separable outputs} if they are structured as}
\begin{equation} \label{eq:outputs}
\small
    \begin{aligned}
        y(x) &= 
        \begin{bmatrix}
            y^r(x) \\ y^s(x_s)
        \end{bmatrix}, \quad
        y^r(x) \in \mathbb{R}^{m_{r}}, \quad
        y^s(x_s) \in \mathbb{R}^{m_{s}},
    \end{aligned}
\end{equation}
\normalsize
\textit{and $y^s(x_s)$ are \textit{separable subsystem outputs} with vector relative degree $\vec{\gamma}^s$. The remaining outputs $y^r(x)$ have vector relative degree $\vec{\gamma}^r$, where $\sum_{i = 1}^{m_r} \gamma^r_i = n_r$, and can depend on any of the system states $x$. The number of subsystem outputs $m_s$ and the number of the rest of the outputs $m_r$ sums to $m$, and $\vec{\gamma} = (\vec{\gamma}^r, \vec{\gamma}^s)$}.

\vp

For the following theorem, we
define the auxilary control input $\mu$ as divided in the following form:
\begin{equation} \label{eq:mu}
    \mu = \begin{bmatrix} \mu_{r}  \\ \mu_{s} \end{bmatrix}, \quad
    \mu_r \in \mathbb{R}^{m_r}, \quad
    \mu_s \in \mathbb{R}^{m_s}.
\end{equation}

\vp

\textbf{\textit{Theorem 1:}} 
\textit{
For a separable control system \eqref{eq:NLform}, if the control law $u(x) = (u_r(x)^T, u_s(x)^T)^T$ \eqref{eq:FeedLin} is constructed with separable outputs \eqref{eq:outputs} and auxiliary control input \eqref{eq:mu}, then $u_s(x) = u_{\ssc}(x)$.
}

\vp

\textit{Proof:} 
We begin by relating the 3 components $(A^{-1}(x),\, L_f^*y(x),\, \mu)$ of $u(x)$ to the components of $u_{\ssc}(x)$, $(A_s^{-1}(x),\, L_{f^{s}}^*y^s(x),\, \mu_s)$. We are given $\mu = \begin{bsmallmatrix} \star \\ \mu_s \end{bsmallmatrix}$ by \eqref{eq:mu}.
With condition \eqref{conditionSSO2}, we show:
\begin{equation*}
\small
    \begin{aligned}
         A(x) &=
        \Bigg(
        \begin{bmatrix}
            \frac{\partial L^{\vec{\gamma}^r - \mathbf{1}}_{f} y^r(x)}{\partial x_r} 
            & \frac{\partial L^{\vec{\gamma}^r - \mathbf{1}}_{f} y^r(x)}{\partial x_s} 
            \\
            \frac{\partial L^{\vec{\gamma}^s - \mathbf{1}}_{f^s} y^s(x)}{\partial x_r} 
            & \dLfy{s}
        \end{bmatrix}
        \begin{bmatrix}
        g^r_1(x) & g^r_2(x) \\
        0     & g^{s}(x) \\
        \end{bmatrix} \Bigg) 
       \\
        &=
        \begin{bmatrix}
            \star & \star
            \\
            0 & \LfLgy{s}
        \end{bmatrix}
        = 
        \begin{bmatrix}
            \star & \star \\
            0 & A_s(x) 
        \end{bmatrix}
        \\
        A(x)^{-1} & =
        \begin{bmatrix}
            \star & \star \\
            0 & A_s(x)^{-1} 
        \end{bmatrix}.
    \end{aligned}
\end{equation*}
Similarly, we show:
\begin{equation*}
\small
    \begin{aligned}
    L_f^*y(x) &=
        \begin{bmatrix}
            \frac{\partial L^{\vec{\gamma}^r - \mathbf{1}}_{f} y^r(x)}{\partial x_r} 
            & \frac{\partial L^{\vec{\gamma}^r - \mathbf{1}}_{f} y^r(x)}{\partial x_s} 
            \\
            \frac{\partial L^{\vec{\gamma}^s - \mathbf{1}}_{f^s} y^s(x)}{\partial x_r} 
            & \dLfy{s}
        \end{bmatrix}
        \begin{bmatrix}
        f^r(x)\\ f^{s}(x)
        \end{bmatrix}
        \\
        &=
        \begin{bmatrix}
            \star \\
            \Lfy{s}
        \end{bmatrix}
        = 
        \begin{bmatrix}
            \star \\
            L_{f^s}^* y^s(x) 
        \end{bmatrix}. 
    \end{aligned}
\end{equation*}
\normalsize
Putting these components together to construct $u(x)$ yields
\begin{equation*}
\small
    \begin{aligned}
        u(x) 
        &= 
        -\begin{bmatrix}
            \star & \star \\
            0 & A_s^{-1}(x) 
        \end{bmatrix}
        \bigg(
        \begin{bmatrix}
            \star \\
            L_{f^s}^* y^s(x) 
        \end{bmatrix} - 
        \begin{bmatrix}
        \star \\ \mu_s
        \end{bmatrix}
        \bigg) 
        \\
        &=
        \begin{bmatrix}
            \star \\
            -A_s^{-1}(x) (L_{f^s}^* y^s(x) - \mu_s) 
        \end{bmatrix} 
        =
        \begin{bmatrix}
            \star \\
            u_{\ssc}(x)
        \end{bmatrix},
    \end{aligned}
\end{equation*}
\normalsize
showing $u_s(x) = u_{\ssc}(x)$ as defined in \eqref{eq:us}. 

\vp

By Theorem 1, we can construct a stabilizing controller \eqref{eq:us} for the subsystem independent of the rest of the system dynamics, identical to the portion of the controller constructed with the full system dynamics acting on the subsystem, therefore guaranteeing full system stability when \eqref{eq:FeedLin} is applied to the rest of the system. This enables construction of stable \textit{model dependent} controllers for separable subsystems without knowledge of the full system dynamics.

\subsection{Equivalency of Subsystems} \label{ssec:Equiv}
Although we now have a subsystem control law independent of the rest of the system dynamics, it still depends on the full system states $x \in \mathbb{R}^n$. Consider the case where the states $x_r \in \mathbb{R}^{n_r}$ cannot be measured. If we could construct an equivalent subsystem whose dynamics are a function of the subsystem states $x_s \in \mathbb{R}^{n_s}$, measurable states $x_r \in \mathbb{R}^{\bar{n}_s}$, and measurable inputs $\mathcal{F} \in \mathbb{R}^{n_f}$, we could calculate the subsystem control law \textit{independent of the full system states}.

Consider another subsystem,
\begin{equation} \label{eq:Subsystem'}
    \begin{aligned}
        &\dot{\bar{x}}_s = \bar{f}^s(\mathcal{X}) + \bar{g}^s(\mathcal{X}) \bar{u}_s
        \\
        & \mathcal{X} = \begin{bmatrix} \bar{x}_r \\ \bar{x}_s \\ \mathcal{F} \end{bmatrix} 
        =\begin{bmatrix} \bar{x}_r \\ x_s \\ \mathcal{F} \end{bmatrix}
        \in \mathbb{R}^{\bar{n}},
    \end{aligned}
\end{equation}
where $\mathcal{X}$ is the state vector $\bar{x} = (\bar{x}_r^T, x_s^T)^T$
augmented with an input $\mathcal{F}$.
Using the same \textit{separable subsystem outputs} as \eqref{eq:outputs_S}, we define a control law $\bar{u}_s(\mathcal{X})$ for the subsystem as:
\begin{equation} \label{eq:us'}
    \begin{aligned}
    \bar{u}_s(\mathcal{X}) &\triangleq 
    -( \underbrace{L_{\bar{g}^s}L_{\bar{f}^s}^{\vec{\gamma}_s - \mathbf{1}} y^s(\mathcal{X})}
    _{\bar{A}_s(\mathcal{X})} )^{-1}
    (\underbrace{L_{\bar{f}^s}^{\vec{\gamma}_s}y^s(\mathcal{X})}
    _{L_{{\bar{f}}^s}^*y^s(\mathcal{X})}
    - \mu_s ) \\
    &= -{\bar{A}}_s^{-1}(\mathcal{X}) (L_{{\bar{f}}^s}^*y^s(\mathcal{X}) - \mu_s).
    \end{aligned} 
\end{equation}

\textit{\textbf{Theorem 2:}}
\textit{
For the subsystems \eqref{eq:Subsystem} and \eqref{eq:Subsystem'},
if $\exists\;
T: \mathbb{R}^{n} \rightarrow \mathbb{R}^{\bar{n}}$
s.t. $T(x) = \mathcal{X}$
and the following conditions hold,}
\begin{equation} \tag{T2} \label{condition1}
    \begin{aligned}
        f^{s}(x) &= \bar{f}^{s}(\mathcal{X}), \\
        g^s(x) &= \bar{g}^{s}(\mathcal{X}),
    \end{aligned}
\end{equation}
\textit{then $u_s(x) = \bar{u}_s(\mathcal{X})$. Applying these to \eqref{eq:Subsystem} and \eqref{eq:Subsystem'}, respectively, results in dynamical systems such that given the same initial condition $\begin{bsmallmatrix} x_{r0} \\ x_{s0} \end{bsmallmatrix} = \begin{bsmallmatrix} x_{r}(t_0) \\ x_{s}(t_0) \end{bsmallmatrix}$ yields solutions $x_s(t) = \bar{x}_s(t)\; \forall t \geq t_0$.}

\vp

\textit{Proof.} 
Since the subsystems have the same dynamics and outputs, the Lie derivatives comprising their control laws are also the same, hence 
\begin{equation*}
    \begin{aligned}
u_s(x) &= -A_s^{-1}(x) (L_{f^{s}}^*y^s(x) - \mu_s) \\
       &= -{\bar{A}}_s^{-1}(\mathcal{X}) (L_{{\bar{f}}^s}^*y^s(\mathcal{X}) - \mu_s) = \bar{u}_s(\mathcal{X}).
    \end{aligned}
\end{equation*}
With the same control law and dynamics, the closed-loop dynamics of the subsystems are the same:
\begin{equation*}
    \begin{aligned}
    \dot{x}_s &= f^s(x) + g^s(x)u_s(x) \\
    &= \bar{f}^s(\mathcal{X}) + \bar{g}^s(\mathcal{X}) \bar{u}_s(\mathcal{X}) = \dot{\bar{x}}_s.
    \end{aligned}
\end{equation*}
Hence, given the same initial condition $\begin{bsmallmatrix} x_{r0} \\ x_{s0} \end{bsmallmatrix} = \begin{bsmallmatrix} x_{r}(t_0) \\ x_{s}(t_0) \end{bsmallmatrix}$, they have the same solution $x_s(t) = \bar{x}_s(t)\; \forall t \geq t_0$.

\vp

By Theorems 1 and 2, we can construct a stabilizing \textit{model dependent} controller, namely \eqref{eq:us'}, for the subsystem \eqref{eq:Subsystem} independent of the rest of the system dynamics and with measurable states and inputs.

\newsec{Zero Dynamics.}
We can apply these theorems to partially feedback linearizable systems by considering the zero dynamics' stability. The full system zero dynamics must either project the same value for the subsystem or not occur in the subsystem.
A formal proof remains for future work. Since we only care about the feedback linearizable part of the system, we focus on full state feedback linearizable systems. 

\section{SEPARABLE ROBOTIC CONTROL SYSTEMS} \label{sec:RobCtrl}
This method of subsystem control for separable systems applies to robotic control systems. While a robotic system may not initially be in the form of \eqref{eq:NLform}, one can construct an equivalent model by dividing the model into 2 subsystems and constraining them to each other through a holonomic constraint \cite{MLS}. For a model in $\dimSpace$-dimensional space, one can consider this holonomic constraint as a $\dof:= \frac{\dimSpace (\dimSpace + 1)}{2}$ DOF fixed joint. Now, the control inputs of one subsystem only affect the other subsystem through the \textit{constraint wrench}. This construction hence decouples the dynamics of one subsystem from the control input of the other so the robotic system can be in separable system form \eqref{eq:NLform}.

\subsection{Robotic System in Separable Form} \label{ssec:RobSep}
Consider an $\drs$ DOF robotic system in $\dimSpace$-dimensional space. 
The coordinates $\theta \in \mathbb{R}^{\drs}$ define the robot's configuration space $\mathcal{Q}$.
Now, consider decomposing this robotic system into 2 subsystems denoted with coordinates $\theta_l \in \mathbb{R}^{\drs_l}$ and $\theta_s \in \mathbb{R}^{\drs_s}$, respectively, and attached with a $\dof$ DOF fixed joint, as previously described, denoted by coordinates $\theta_f$, where $\drs_l + \drs_s + \dof = \eta$. An example of this configuration for an amputee-prosthesis system is shown in Fig. 2.

The dynamics of a robotic system are given by the classical Euler-Lagrangian equation \cite{ModelsGrizzle}, \cite{MLS}:
\begin{equation} \label{eq:robotDyn}
D(\theta) \ddot{\theta} + H(\theta, \dot{\theta}) = B u + J^T(\theta) F.
\end{equation}
\noindent Here $D(\theta) \in \mathbb{R}^{\drs \times \drs}$ is the inertial matrix. $H(\theta, \dot{\theta}) = C(\theta, \dot{\theta}) + G(\theta) \in \mathbb{R}^{\drs}$, a vector of centrifugal and Coriolis forces and a vector containing gravity forces, respectively. The actuation matrix $B \in \mathbb{R}^{\drs\times m}$ contains the gear-reduction ratio of the actuated joints and is multiplied by the control inputs $u \in \mathbb{R}^{m}$, where $m$ denotes the number of control inputs. The wrenches $F \in \mathbb{R}^{\dhl_h}$ enforce the $\dhl_h$ holonomic constraints. The Jacobian matrix of the holonomic constraints $J(\theta) = \frac{\partial h}{\partial \theta} \in \mathbb{R}^{\dhl_h \times \drs}$ enforces the holonomic constraints by:
\begin{equation}\label{eq:holoConstr}
\dot{J}(\theta, \dot{\theta}) \dot{\theta} + J(\theta) \ddot{\theta} = 0.
\end{equation}
Solving \eqref{eq:robotDyn} and \eqref{eq:holoConstr} simultaneously yields the \textit{constrained dynamics}. These terms will now be referred to as $D,\, H,\, \text{and } J$, respectively, for notational simplicity.

Here $\dhl_h$ is the summation of the number of holonomic constraints for each subsystem, $\dhl_{h, l}$ and $\dhl_{h, s}$ respectively, and the fixed joint $\dof$, i.e. $\dhl_h = \dhl_{h, l} + \dhl_{h, s} + \dof$. The wrench $F$ includes the fixed joint wrenches $F_f \in \mathbb{R}^{\dof}$, i.e. $F_f \subset F$, and the Jacobian $J$ includes the Jacobian of the fixed joint's holonomic constraints $J_f = \frac{\partial h_f}{\partial \theta}\in \mathbb{R}^{\dof \times \drs}$, i.e $J_f \subset J$. 

\begin{figure} 
    \centering
    \includegraphics[width=0.27\textwidth]{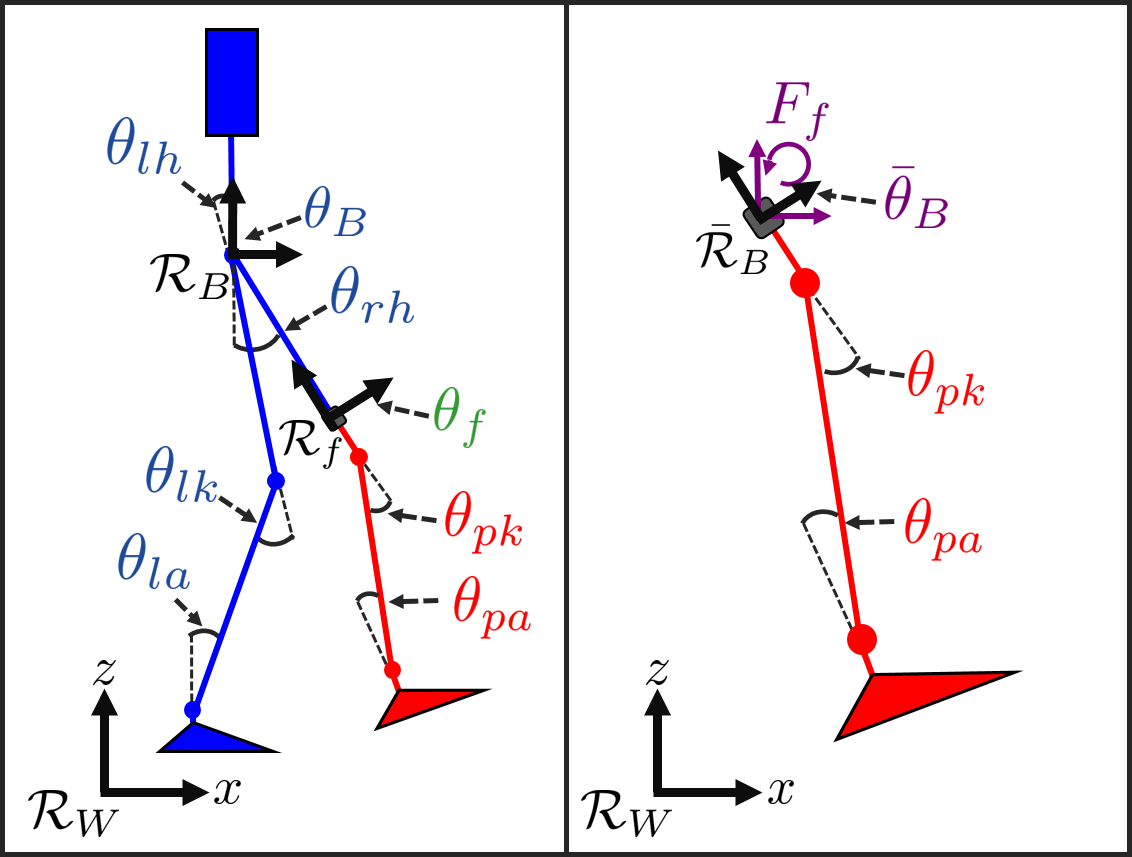}  \label{fig:FullSystCoord}
    \vspace{-0.3cm}
    {\caption{(Left) Model of the full amputee-prosthesis system labeled with its generalized coordinates, where $\theta_l$ is notated in blue, $\theta_f$ in green, and $\theta_s$ in red. (Right) Model of prosthesis subsystem with its generalized coordinates, where $\bar{\theta}_B$ and $F_f$ are notated in violet.}}
    \vspace{-0.8cm}
\end{figure}

\newsec{Robotic System in Nonlinear Form.}
Using the notation from Section \ref{ssec:CtrlSep}, $x_s = (\theta_s^T, \dot{\theta}_s^T)^T \in \mathbb{R}^{n_s}$ will denote the states of the subsystem under consideration for control, and $x_r = (\theta_l^T, \theta_f^T, \dot{\theta}_l^T, \dot{\theta}_f^T)^T = (\theta_r^T, \dot{\theta}_r^T)^T  \in \mathbb{R}^{n_r}$ will denote the states of the rest of the system. 
We also define the control input $u = (u_r^T, u_s^T)^T$, where $u_r \in \mathbb{R}^{m_r}$ and $u_s \in \mathbb{R}^{m_s}$. By defining the following functions,
\begin{equation} \label{eq:nlDyn}
    \begin{aligned}
        &\begin{bmatrix}
            \hat{f}^r(\theta, \dot{\theta})_{(\drs_l + \dof) \times 1} \\ 
            \hat{f}^s(\theta, \dot{\theta})_{\drs_s \times 1}
        \end{bmatrix}
       \triangleq
        D^{-1} (- H + J^T F), 
        \\
        &\begin{bmatrix}
            \hat{g}^r_1(\theta, \dot{\theta})_{(\drs_l + \dof) \times m_r} &
            \hat{g}^r_2(\theta, \dot{\theta})_{(\drs_l + \dof) \times m_s} \\ 
            \hat{g}^r_3(\theta, \dot{\theta})_{\drs_s \times m_r} &
            \hat{g}^s(\theta, \dot{\theta})_{\drs_s \times m_s} 
        \end{bmatrix}
        \triangleq
        D^{-1} B,
    \end{aligned}
\end{equation}
we construct the robotic system in nonlinear form with $x = (x_r^T, x_s^T)^T$:
\begin{equation} \label{eq:RobSyst}
    \begin{aligned}
        \dot{x} &= f(x) + g(x)u \\
        &= 
        \underbrace{
        \begin{bmatrix}
        \dot{\theta}_r \\
        \hat{f}^r(\theta, \dot{\theta}) \\ 
        \dot{\theta}_s \\
        \hat{f}^s(\theta, \dot{\theta}) \\ 
        \end{bmatrix}
        }_{f(x)}
        +
        \underbrace{
        \begin{bmatrix}
        0 &
        0 \\
        \hat{g}^r_1(\theta, \dot{\theta}) & 
        \hat{g}^r_2(\theta, \dot{\theta}) \\ 
        0 &
        0 \\
        \hat{g}^r_3(\theta, \dot{\theta}) & 
        \hat{g}^s(\theta, \dot{\theta}) \\ 
        \end{bmatrix}
        }_{g(x)}
        \begin{bmatrix}
        u_r \\ u_s
        \end{bmatrix}
    \end{aligned}
\end{equation}
To consider the separability of this system, we examine an equivalent representation of this system with the fixed joints coordinates $\theta_f$ in the world frame $\mathcal{R}_W$ instead of the fixed joint reference frame $\mathcal{R}_f$. This yields a block diagonal inertia matrix $\tilde{D}(\theta)$ \cite[p. 280]{MLS} with block dimensions $\eta_l \times \eta_l$ and $(\delta + \eta_s) \times (\delta + \eta_s)$ and hence the inverse is a block diagonal matrix with the same block dimensions. 
We note the lower left $(\delta + \eta_s) \times \eta_l$ block of $\tilde{D}^{-1}(\theta)$ is 0.

Our system's inertia matrix $D(\theta)$ relates to $\tilde{D}(\theta)$ by $D = \tilde{J}^{T} \tilde{D} \tilde{J}$, where $\tilde{J}(\theta) \in \mathbb{R}^{\eta \times \eta}$ is the Jacobian of the forward kinematics map between the original coordinates and the transformed coordinates \cite[pp. 283-285]{MLS}, 
\begin{equation} \label{eq:JacobianTransform}
    \tilde{J}(\theta) =
    \begin{bmatrix}
    I & 0 & 0 
    \\
    \frac{\partial g_{Wf}(\theta)}{\partial \theta_l} 
    &
    \frac{\partial g_{Wf}(\theta)}{\partial \theta_f}
    &
    \frac{\partial g_{Wf}(\theta)}{\partial \theta_s} 
    \\
    0 & 0 & I
    \end{bmatrix}.
\end{equation}
Here $g_{Wf}(\theta)$ is the forward kinematics of $\mathcal{R}_f$ with respect to $\mathcal{R}_W$. 
The form of \eqref{eq:JacobianTransform} yields $\tilde{J}^{-1}$ and $\tilde{J}^{-T}$ as matrices with non-identity components only in the columns corresponding to $\theta_f$. Hence when relating the inverse inertia matrix by $D^{-1} = \tilde{J}^{-1} \tilde{D}^{-1} \tilde{J}^{-T}$, only components relating to the fixed joint are transformed so the lower left block $0_{(\delta + \eta_s) \times \eta_l}$ is preserved. 
Since the fixed joints are unactuated and the subsystem joints are only actuated by $u_s$, $B$ has a lower left block $0_{(\delta+\eta_s) \times m_s}$. Therefore $D^{-1} B$ yields a lower left block $0_{\eta_s \times m_r}$ in $g(x)$, i.e. $\hat{g}_3^r(\theta, \dot{\theta}) = 0$.
We numerically verified this result for the amputee-prosthesis system in this paper. More details of this proof will be included in a future work.

We can now write \eqref{eq:RobSyst} as a \textit{separable control system} \eqref{eq:NLform}, 
\begin{equation} \label{eq:robSepSyst}
\small
    \begin{aligned}
        \begin{bmatrix}
        \dot{x}_r \\ \dot{x}_s
        \end{bmatrix}
        &= 
        \begin{bmatrix}
            \begin{bmatrix}
                \dot{\theta}_r \\
                \hat{f}^r(\theta, \dot{\theta})
            \end{bmatrix}
        \\
            \begin{bmatrix}
                \dot{\theta}_s \\
                \hat{f}^s(\theta, \dot{\theta})
            \end{bmatrix}
        \\ 
        \end{bmatrix}
        +
        \begin{bmatrix}
            \begin{bmatrix}
                0\\
                \hat{g}^r_1(\theta, \dot{\theta})
            \end{bmatrix}
            &
            \begin{bmatrix}
            0 \\
            \hat{g}^r_1(\theta, \dot{\theta})
            \end{bmatrix}
            \\
            \begin{bmatrix}
                0 \\
                0
            \end{bmatrix}
            &
            \begin{bmatrix}
            0 \\
            \hat{g}^s(\theta, \dot{\theta})
            \end{bmatrix}
       \end{bmatrix}
        \begin{bmatrix}
        u_r \\ u_s
        \end{bmatrix} 
        \\
        & \triangleq
        \fmatrix{}
        +
        \gmatrix{} 
        \begin{bmatrix}
        u_r \\ u_s 
        \end{bmatrix},
    \end{aligned}
\end{equation}
\normalsize
which implies a \textit{separable subsystem} can be written as \eqref{eq:Subsystem}.

\noindent \textbf{Remark 2.} 
By Theorem 1, for any \textit{separable subsystem outputs} $y^s(x_s)$, a feedback linearizing control law $u_s(x)$ \eqref{eq:us} can be constructed.
Typically, we calculate $u(x)$ with $F$ in terms of $u$ by solving \eqref{eq:robotDyn} for $\ddot{\theta}$ and substituting it into \eqref{eq:holoConstr}. Solving for $F$ yields
\begin{equation} \label{eq:force}
    \begin{aligned}
    F(\theta, \dot{\theta}, u) &= (J D^{-1} J^T)^{-1} (J D^{-1}(H - Bu) - \dot{J} \dot{\theta})
    \\
    &\triangleq
    \lambda_f + \lambda_g u.
    \end{aligned}
\end{equation}
Then we rearrange the dynamics in \eqref{eq:nlDyn} as $D^{-1}(-H + J^T \lambda_f)$ and $D^{-1}(B + J^T \lambda_g$), respectively.  While this arrangement does not yield a separable control system, 
it calculates the same $u(x)$, hence Theorem 1 still applies.

\subsection{Equivalent Robotic Subsystem and Controller} \label{ssec:EquivRob}
The advantage of constructing a robotic system in the form \eqref{eq:NLform} is applying Theorem 2 to construct an equivalent control law for a robotic subsystem without knowledge of the full system dynamics and states.

\noindent \textbf{Construction of Equivalent Robotic Subsystem. }
We construct an equivalent robotic subsystem, Fig. 2, by modeling the links and joints of the full model denoted by coordinates $\theta_s$. We place a $\dof$ DOF base frame, denoted by $\bar{\theta}_B$, at the end of the link connected to the fixed joint in the full model. 
At these base coordinates we project interaction forces and moments $F_f$ with $\bar{J}_{f}^T = (R_{Wf}(\bar{\theta}_B) [I_{\dof \times \dof} \; 0_{\dof \times n_s}])^T$. Here $R_{Wf}^T(\bar{\theta}_B)$ rotates $F_f$ from $\mathcal{R}_{f}$ to $\mathcal{R}_{W}$.
With configuration coordinates $\bar{\theta} = (\bar{\theta}_B; \theta_s)$, the \textit{constrained subsystem dynamics} equations are
\begin{align} \label{eq:robotDyn'}
    \bar{D}(\bar{\theta}) \ddot{\bar{\theta}} + \bar{H}(\bar{\theta}, \dot{\bar{\theta}}) &= \bar{B} \bar{u} + \bar{J}^T(\bar{\theta}) \bar{F}(\bar{\theta}, \dot{\bar{\theta}}) + \bar{J}_{f}^T(\bar{\theta}) F_f,
    \\ \label{eq:holoConstr'}
    \dot{\bar{J}} \dot{\bar{\theta}} + \bar{J} \ddot{\bar{\theta}} &= 0,
\end{align}
where $\bar{J}(\bar{\theta}) = \frac{\partial \bar{h}}{\partial \bar{\theta}} \in \mathbb{R}^{\eta_{h, s} \times (\dof + \drs_s)}$ is the Jacobian of the holonomic constraints acting on the subsytem, apart from those for the fixed joint. 
These terms will now be referred to as $\bar{D},\, \bar{H},\, \bar{J}, \bar{J}_f,\, \text{and } \bar{F}$, respectively.

\newsec{Transformation for Subsystem States and Inputs.}
Using the notation from Section \ref{ssec:Equiv}, with $x_s$ from the full system, $\bar{x}_r = (\bar{\theta}_B^T, \dot{\bar{\theta}}_B^T)^T \in \mathbb{R}^{\bar{n}_r}$, and $\mathcal{F} = F_f \in \mathbb{R}^{n_f}$, we construct the robotic subsystem in nonlinear form. We relate the subsystem's augmented state vector $\mathcal{X}$ to the full system states $x$. 
The base coordinates $\bar{\theta}_B$ relate to the full system through the forward kinematics,
$\bar{\theta}_B = g_{Wf}(\theta)$ \cite{MLS}. Their velocities $\dot{\bar{\theta}}_B$ relate by the \textit{spatial manipulator Jacobian}, $\dot{\bar{\theta}}_B = J_{Wf}^s(\theta)\dot{\theta}$ \cite{MLS}. To obtain an expression for $F_f$ based on the equation for $F(\theta, \dot{\theta}, u)$ given in \eqref{eq:force}, we define the transformation $\iota: \mathbb{R}^{\dhl_h} \rightarrow \mathbb{R}^\dof$ where $\iota_f(F) = \iota_f(F \supset F_f) = F_f$. Hence the transformation $T(x) = \mathcal{X}$ is defined as:
\begin{equation*}
    T(x) \triangleq 
    \begin{bmatrix} 
    g_{Wf}(\theta) \\ 
    J_{Wf}^s(\theta)\dot{\theta} \\
    x_s \\
    \iota_f(F(\theta, \dot{\theta}, u))
    \end{bmatrix}
    = 
    \begin{bmatrix}
    \bar{\theta}_B \\ 
    \dot{\bar{\theta}}_B \\
    x_s \\
    F_f
    \end{bmatrix}
    = \mathcal{X}.
\end{equation*}

\newsec{Robotic Subsystem in Nonlinear Form.}
By defining the following functions,
\begin{equation} \label{eq:nlDyn'}
    \begin{aligned}
        \begin{bmatrix}
            \hat{\bar{f}}^r(\bar{\theta}, \dot{\bar{\theta}}, F_f)_{\dof \times 1} \\ 
            \hat{\bar{f}}^s(\bar{\theta}, \dot{\bar{\theta}}, F_f)_{{\drs}_s \times 1}
        \end{bmatrix}
        &\triangleq
        \bar{D}^{-1} (- \bar{H} + \bar{J}^T \bar{F} + \bar{J}_{f}^T F_f),
        \\
        \begin{bmatrix}
            \hat{\bar{g}}^r(\bar{\theta}, \dot{\bar{\theta}})_{\dof \times m_s} \\ 
            \hat{\bar{g}}^s(\bar{\theta}, \dot{\bar{\theta}})_{{\drs}_s \times m_s}
        \end{bmatrix}
       &\triangleq
        \bar{D}^{-1} \bar{B},
    \end{aligned}
\end{equation}
we construct the equivalent robotic subsystem as \eqref{eq:Subsystem'}:
\begin{equation} \label{eq:robSepSyst'}
    \begin{aligned}
        \dot{\bar{x}}_s
        &= 
        \begin{bmatrix}
            \dot{\theta}_s \\
            \hat{\bar{f}}^s(\bar{\theta}, \dot{\bar{\theta}}, F_f)
        \end{bmatrix}
        +
        \begin{bmatrix}
            0_{\drs_s \times m_s} \\
            \hat{\bar{g}}^s(\bar{\theta}, \dot{\bar{\theta}})
        \end{bmatrix}
        \bar{u}_s \\
        & \triangleq
        \bar{f}^s(\mathcal{X}) + \bar{g}^s(\mathcal{X}) \bar{u}_s.
    \end{aligned}
\end{equation}
To examine the subsystem equivalency, we again consider the system with $\theta_f$ in world coordinates where the transformed system is (11) right-multiplied by $\tilde{J}^{-T}$ with $D = \tilde{J}^T \tilde{D} \tilde{J}$. Here the lower block of $\tilde{D}$ and the bottom $\delta + \eta_s$ components of the rest of dynamics give exactly (18). Since $\tilde{J}^{-T}$ only affects components relating to the fixed joint coordinates, $x_s = (\dot{q}_s^T, \ddot{q}_s^T)^T$ is invariant to the transformation, such that condition (T2) is met for the given structure of the robotic subsystem model. We numerically verified condition \eqref{condition1} is met for the amputee-prosthesis system in this paper. More details of this proof will be included in a future work.

\noindent \textbf{Remark 3.} 
With condition \eqref{condition1} met, Theorem 2 applies, 
enabling users to construct controllers for a robotic subsystem, without knowledge of the full dynamics and states, given the constraint forces and moments $F_f$ at the fixed joint and its global coordinates $\bar{\theta}_B$ and velocities $\dot{\bar{\theta}}_B$. This control law yields the same evolution of subsystem states $x_s$ as those of the full system under the control law $u(x)$.
As explained for the full system, we typically solve $\bar{F}$ in terms of $\bar{u}_s$ and calculate $\bar{u}_s$ with rearranged dynamics in \eqref{eq:nlDyn'}. While this form does not meet \eqref{condition1}, Theorem 2 still holds since the method calculates the same $\bar{u}_s$.

To support our arguments for robotic system separability and subsystem equivalency with experimental data, we examined human-prosthesis motion capture walking data \cite{gehlhar2020data} and computed the control input with inverse dynamics.
With position and time data, we computed discrete accelerations. By averaging these over a time window we obtain accelerations for computing the inverse dynamics with the full system dynamics \eqref{eq:robotDyn} and equivalent subsystem dynamics \eqref{eq:robotDyn'}. This yielded identical prosthesis control inputs. See Fig. 1.

\section{AMPUTEE-PROSTHESIS APPLICATION} \label{sec:AmpProsth}
The control methods presented in \ref{ssec:CtrlSep} and \ref{ssec:Equiv} apply to an amputee-prosthesis system 
by modeling it as a 2D bipedal robot per the methods of \cite{ModelsGrizzle} with the addition of a fixed joint at the socket per the methods in \ref{ssec:RobSep}. The prosthesis subsystem is constructed according to the method in \ref{ssec:EquivRob}.
The parameters for the human limbs are estimated based on the user's total height and mass. The length, mass, and COM are calculated based on Plagenhoef's table of percentages \cite{HumanParam}. The inertia of each limb is calculated based on Erdmann's table of radiuses of gyration \cite{HumanInertia}. The prosthesis parameters are based on a CAD model of AMPRO3 \cite{zhao2017preliminary}, a powered transfemoral prosthesis, Fig. 1.

\newsec{Generalized Coordinates.}
For the generalized coordinates $\theta = (\theta_l^T,\, \theta_f^T,\, \theta_{s}^T)^T$ of \ref{sec:RobCtrl}, we define $\theta_l$ as $(\theta_B^T,\, \theta_{ll}^T,\, \theta_{rl}^T)^T$
Here $\theta_B \in SE(2)$ are the extended coordinates representing position and rotation of the full system's base frame $\mathcal{R}_{B}$ with respect to the world frame $\mathcal{R}_{W}$, and $\theta_{ll} = (\theta_{lh}^T,\, \theta_{lk}^T,\, \theta_{la}^T)^T$ and $\theta_{rl} = \theta_{rh}$ are the relative joint angles of the human's left and right leg, respectively. 
The subsystem coordinates $\theta_{s} = (\theta_{pk}^T,\, \theta_{pa}^T)^T$ denote the relative joint angles of the prosthesis knee and ankle, respectively.
See Fig. 2.

\subsection{Hybrid Systems} \label{ssec:HybridSyst}
To model human locomotion which consists of both continuous and discrete dynamics, a hybrid system is employed.
This \textit{multi-domain hybrid control system} is defined as a tuple,
\begin{equation*}
    \HybridControlSystem = (\DirectedGraph,\, \Domain,\, \ControlInput,\, \Guard,\, \Delta,\, FG),
\end{equation*}
\noindent where:
where $\DirectedGraph = (V,\, E)$ is a \textit{directed cycle}, with vertices $V = \{v_1 = \pt,\, v_2 = \pw\}$ and edges $E = \{e_1 = \{\pt \rightarrow \pw\}, e_2 = \{\pt \rightarrow \pw\}\}$, where \textit{\pt} represents ``\textbf{p}rosthesis s\textbf{t}ance", and \textit{\pw} ``\textbf{p}rosthesis s\textbf{w}ing". The set of domains of admissibility is given by $\Domain = \{\Domain_{v}\}_{v \in V}$, and set of admissible control inputs by $\ControlInput = \{\ControlInput_{v}\}_{v \in V}$. A domain ends when the system reaches a guard in the set $\Guard = \{\Guard_{e}\}_{e \in E}$, which triggers the discrete transitions defined by the set of reset maps $\Delta = \{\Delta_{e}\}_{e \in E}$. The continuous dynamics $\dot{x} = f_v(x) + g_v(x)u_v$ for each domain are defined by $(f_v,\, g_v)$ in the set of control systems $FG = \{(f_v,\, g_v)\}_{v \in V}$. Details in \cite{ames2014human}. See Fig. 3 for a depiction of the directed cycle.

\begin{figure} \label{fig:DirectedCycle}
\centering
\includegraphics[width=0.3\textwidth]{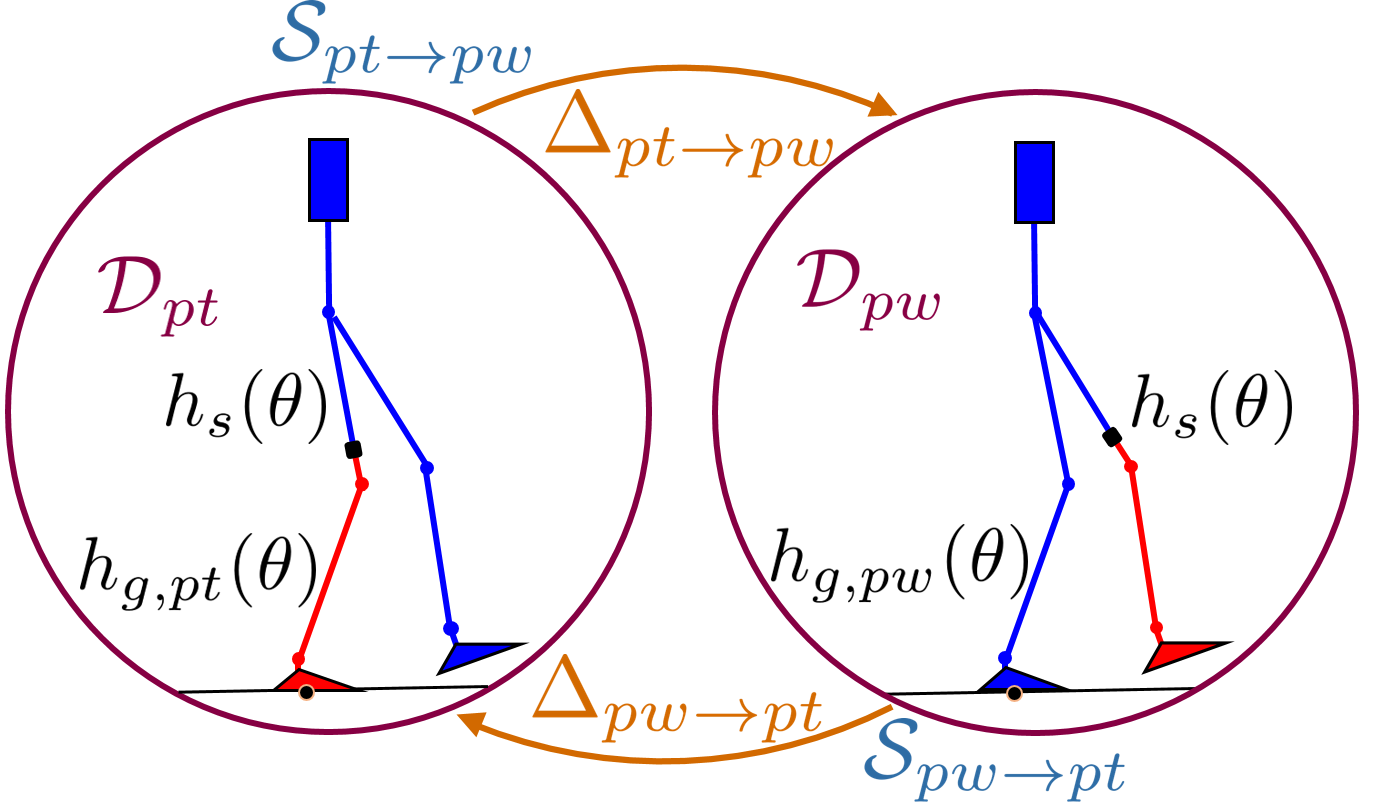}
\vspace{-0.4cm}
{\caption{The walking pattern of an asymmetric amputee-prosthesis system can be represented as a 2-step \textit{directed cycle}.}}
\vspace{-0.7cm}
\end{figure} 

\noindent \textbf{Continuous Domains of Full System.}
The dynamics of the full system are given by \eqref{eq:robotDyn}, and control system by \eqref{eq:RobSyst}. 
For each domain, a 3 DOF holonomic constraint $h_{g, v}(\theta)$ is used to model the foot contact points present in human walking behavior \cite{ModelsGrizzle}. Both domains also have a holonomic constraint $h_{s}(\theta)$ for the 3 DOF fixed joint at the socket.
See Fig. 3.
For this system, $m_s = 2$ for the number of actuated joints in the prosthesis subsystem $(\theta_{pk},\, \theta_{pa})$, and $m_r = 4$ for the rest of the of actuated joints $(\theta_{lh},\, \theta_{lk},\, \theta_{la}\, \theta_{rh})$. 

\noindent \textbf{Continuous Domains for Prosthesis Subsystem.}
The continuous dynamics for both domains are defined by \eqref{eq:robotDyn'} where 
$\bar{J}^T$ is only present for \textit{prosthesis stance} when $\bar{h}_{g, \pt}$ defines the prosthesis foot contact. The control system is defined by \eqref{eq:robSepSyst'}.
As explained in \ref{ssec:Equiv} and \ref{ssec:EquivRob}, the prosthesis subsystem must measure its states $\bar{x}_r = (\bar{\theta}_B^T, \dot{\bar{\theta}}_B^T)^T$ and input $\mathcal{F} = F_f$.
In $\Domain_{\pt}$, $\bar{\theta}_B$ and $\dot{\bar{\theta}}_B$ can be calculated using inverse kinematics of its own joints since the foot is assumed to be flat on the ground. In $\Domain_{\pw}$, an IMU on the amputee's amputated thigh can provide the global rotation and velocities. The global positions are not required since they do not appear in the dynamics. Both domains require a load cell at the socket interface to measure the interaction forces $F_f$.

\noindent \textbf{Guards and Reset Maps.}
For both domains, the condition for the guard $\Guard$ is met when the swing leg reaches the ground. The reset map $\Delta$ dictates the discrete dynamics that occur assuming a perfectly plastic impact. See \cite{RigidCollision} for details.

\subsection{Prosthesis Gait Design} \label{ssec:GaitGen}
To control the prosthesis to yield stable human-like walking for the entire amputee-prosthesis system, we design desired walking trajectories $y^d$ for the actual motion $y^a$ of the robotic system to follow. We encode the motion through outputs for the full system, 
\begin{equation*}
\begin{aligned}
y_{1,v}(\theta, \dot{\theta}) &= y_{1,v}^a(\theta, \dot{\theta}) - y_{1, v}^d, \\
y_{2,v}(\theta) &= y_{2,v}^a(\theta) - y_{2,v}^d(\tau_v, \alpha_v),
\end{aligned}
\end{equation*}
for $v \in V$, where $y_{1,v}$ and $y_{2,v}$ are relative degree 1 and 2 outputs, respectively \cite{sastry2013nonlinear}. The feedback linearizing controller \eqref{eq:FeedLin} constructed with these outputs will drive these outputs to 0 such that $y^a = y^d$.
Here $y_{1, v}^d$ is defined as the forward hip velocity $v_{hip, v}$, which appeared approximately constant in human locomotion data \cite{ames2014human}, and $y_{2, v}^d(\tau_v, \alpha_v)$ is defined as a $6^{th}$ order B\'ezier polynomial. The values $v_{hip, v}$ and $\alpha_v$ are determined through an optimization method we will describe. Here $\tau_v$ is a phase variable that measures the forward progression of the walking trajectory and enables development of a robust state-feedback controller \cite{Westervelt07feedbackcontrol}.

\noindent \textbf{Partial Hybrid Zero Dynamics and Optimization.}
The controller \eqref{eq:FeedLin} 
drives the outputs $y_{1, v}$ and $y_{2, v}$ to $0$ over the continuous dynamics of the domain, rendering invariant zero dynamics.
However, the zero dynamics do not necessarily remain invariant through impacts. 
In fact, enforcing impact invariance on the velocity-modulating output is too restrictive due to the jump of velocities by the impact map.
Hence, we only enforce an impact invariance condition on the relative degree 2 outputs, resulting in \emph{partial zero dynamics}:
\begin{equation*}
\small
\textbf{PZ}_{\alpha_v} = \{(\theta, \dot{\theta}) \in \mathcal{D}_v: \, y_{2, v}(\theta, \alpha_v) = 0,\, \dot{y}_{2, v}(\theta, \dot{\theta}, \alpha_v) = 0 \}.
\end{equation*}
We design the trajectories with \textit{partial hybrid zero dynamics} (PHZD) constraints for each transistion,
\begin{align*}
    \Delta_{e_i}(\Guard_{e_i} \cap \textbf{PZ}_{\alpha_{v_i}}) \subseteq \textbf{PZ}_{\alpha_{v_{i + 1}}},
\end{align*}
in a direct collocation based multi-domain HZD gait optimization approach, as described in \cite{hereid20163d}. Here $v_{i + 1}$ indicates the next domain in the directed cycle. The solution gives the sets of $v_{hip, v}$ and $\alpha_v$ for each domain $\Domain_v$, defining the desired output functions $y^d_v$ to yield stable amputee-prosthesis human-like walking.

\noindent \textbf{Phase Variable.}
Previous work found the forward progression of the stance hip $p_{hip}$ to be monotonic during a human step cycle \cite{ames2014human}, qualifying it to be a phase variable for walking trajectories. We linearize this expression as $\delta p_{hip}(\theta)$ and formulate it as a phase variable:
\begin{equation}\label{eq:tau}
\tau_v(x) = \frac{\delta p_{hip, v}(\theta) - \delta p_{hip, v}^+}{\delta p_{hip, v}^- - \delta p_{hip, v}^+}.
\end{equation}
\noindent Here $\delta p_{hip, v}^+$ and $\delta p_{hip, v}^-$ are the starting and ending positions in a step, respectively, determined through the optimization method. In  $\Domain_{\pt}$, 
$\tau_{\pt}$ is a function of the prosthesis states $x_s$, but in $\Domain_{\pw}$, it needs IMUs on the human's stance leg to measure the phase variable in real-time. It can differentiate this signal to calculate the time derivatives $\dot{\tau}_{\pw}$ and $\ddot{\tau}_{\pw}$.

\newsec{Separable Outputs.}
To design \textit{separable outputs} to enable separable subsystem control, we define the actual outputs $y^a_v$:
\begin{itemize}
    \item{\makebox[3cm][l]{hip velocity:} $y_{1, v}^{a, \rm{vhip}} = r^{\rm{sk}}_v \dot{\theta}^{\rm{sk}}_v + (r^{\rm{sk}}_v+ r^{\rm{sa}}_v) \dot{\theta}^{\rm{sa}}_v$} 
    \item{\makebox[3cm][l]{stance calf:} $y_{2, v}^{a, \rm{sc}} \,\,\,\,\, = -\theta^{\rm{sk}}_v - \theta^{\rm{sa}}_v$,} 
    \item{\makebox[3cm][l]{stance hip:} $y_{2, v}^{a, \rm{sh}} \,\,\,\,\, = -\theta^{\rm{sh}}_v$,} 
    \item{\makebox[3cm][l]{non-stance hip:} $y_{2, v}^{a, \rm{nsh}} \,\, = -\theta^{\rm{nsh}}_v$,} 
    \item{\makebox[3cm][l]{non-stance knee:} $y_{2, v}^{a, \rm{nsk}} \,\, = \theta^{\rm{nsk}}_v$,} 
    \item{\makebox[3cm][l]{non-stance ankle:} $y_{2, v}^{a, \rm{nsa}} \,\, = \theta^{\rm{nsa}}_v$.} 
\end{itemize}
Here $r^{\rm{sk}}_v$ and $r^{\rm{sa}}_v$ are the stance knee and ankle limb lengths, respectively. The angles $(\theta^{\rm{sh}}_v,\, \theta^{\rm{sk}}_v,\, \theta^{\rm{sa}}_v,\, \theta^{\rm{nsh}}_v,\, \theta^{\rm{nsk}}_v,\, \theta^{\rm{nsa}}_v)$ are defined as $(\theta_{rl},\, \theta_{s},\, \theta_{ll})$ for $\Domain_{\pt}$ and $(\theta_{ll},\, \theta_{rl},\, \theta_{s})$ for $\Domain_{\pw}$. The \textit{separable subsystem outputs} are defined as $y^s_{\pt}(x_s) = (y_{1, \pt}^{\rm{vhip}},\, y_{2, \pt}^{\rm{sc}})$ for $\Domain_{\pt}$ and $y^s_{\pw}(x_s) = (y_{2, \pw}^{\rm{nsk}},\, y_{2, \pw}^{\rm{nsa}})$ for $\Domain_{\pw}$. 
In $\Domain_{\pt}$, we have $\tau_{\pt}(x_s)$ and hence $y_{1, \pt}^{\rm{vhip}}$ satisfies \eqref{conditionSSO2} because clearly $\frac{\partial y_1^s(x_s)}{\partial x_r} = 0$ and \eqref{conditionSSO1} does not apply since its relative degree $\gamma^s_1$ is 1. Because the relative degree 2 outputs are based on position and also the signal $\tau_{\pw}$ in $\Domain_{\pw}$, their Lie derivatives are the velocities of the specified prosthesis joints, meaning $\frac{\partial L_{f^s}y^s_2(x)}{\partial x_r} = 0$, satisfying \eqref{conditionSSO1} and \eqref{conditionSSO2}.
The remaining outputs for each domain define $y_{r, \pt}(x)$ and $y_{r, \pw}(x)$, respectively. 

\subsection{Prosthesis State-based Control} \label{ssecApplication}
Since $\tau_{\pt}$ is a function of $x$, $x$ is the only time dependent variable in the output functions $y_{\pt}(x)$, meaning a control law for the full system 
$u_{\pt}(x) = ( u_{r, \pw}(x)^T,  u_{s, \pt}(x)^T )^T$
can be defined by \eqref{eq:FeedLin} and for the subsystem $\bar{u}_{s, \pt}(\mathcal{X})$ by \eqref{eq:us'}. By Theorems 1 and 2, $u_{s, \pt}(x) = \bar{u}_{s, \pt}(\mathcal{X})$.

For $\Domain_{\pw}$, $\tau_{\pw}$ is measured and hence not a function of $x$, so we define the control laws slightly differently
to account for the time dependency of $\tau_{\pw}$,
\begin{equation} \label{eq:FeedLinT}
\resizebox{0.89\hsize}{!}{$
    u_{\pw}(x, \mathcal{T}) = 
    \begin{bmatrix}
        u_{r, \pw}(x, \mathcal{T}) \\
        u_{s, \pw}(x, \mathcal{T})
    \end{bmatrix}
    = -A(x)^{-1} (L_f^* y(x) - 
    \begin{bmatrix}
    \dot{y}_{1, \pw}^d \\ 
    \ddot{y}_{2, \pw}^d
    \end{bmatrix}
    - \mu),
    $}
\end{equation}
where $\mathcal{T} = (\tau_{\pw}, \dot{\tau}_{\pw}, \ddot{\tau}_{\pw})$, $A(x)$ and $L_f^*y(x)$ from \eqref{eq:FeedLin}, and
\begin{equation*}
        \begin{bmatrix}
    \dot{y}_{1, \pw}^d \\ 
    \ddot{y}_{2, \pw}^d
    \end{bmatrix}
    = 
    \begin{bmatrix}
    \frac{\partial y_{1, \pw}^d(\tau_{\pw}, \alpha_{\pw})}{\partial \tau_{\pw}} \dot{\tau}_{\pw} \\
    \frac{\partial^2 y_{2, \pw}^d(\tau_{\pw}, \alpha_{\pw})}{\partial \tau^2_{\pw}}\dot{\tau}_{\pw}^2 + \frac{\partial y_{2, \pw}^d(\tau_{\pw}, \alpha_{\pw})}{\partial \tau_{\pw}} \ddot{\tau}_{\pw}
    \end{bmatrix}.
\end{equation*}
Note the Lie derivatives of $y_{\pw}(x)$ result in only being with respect to $y^a_{\pw}(x)$ since $y^d_{\pw}(\tau, \alpha)$ is not a function of $x$.
A feedback linearizing controller for the subsystem is given by
\begin{equation} \label{eq:usT}
        \bar{u}_{s, \pw}(\mathcal{X}, \mathcal{T}) = - A_{s}^{-1}(\mathcal{X}) (L_{f^s}^* y^{s}(\mathcal{X}) - \ddot{y}^{s, d}_{\pw} - \mu_s),
\end{equation}
where $A_s(\mathcal{X})$ and $L_f^*y(\mathcal{X})$ are defined as in \eqref{eq:us'} and
\begin{equation*}
    \ddot{y}^{s, d}_{2, \pw} = \frac{\partial^2 y_{2, \pw}^{s, d}(\tau_{\pw}, \alpha_{\pw})}{\partial \tau^2_{\pw}}\dot{\tau}^2_{\pw} + \frac{\partial y_{2, \pw}^{s, d}(\tau_{\pw}, \alpha_{\pw})}{\partial \tau_{\pw}} \ddot{\tau}_{\pw}.
\end{equation*}
Here also, the Lie derivatives of $y^s_{\pw}(x_s)$ result in only being with respect to $y^{s, a}_{\pw}$ since $y^{s, d}_{\pw}$ is not a function of $x$.

\vp

\textit{\textbf{Proposition 1. }}
\textit{
For a separable control system \eqref{eq:NLform} with separable outputs \eqref{eq:outputs}, if the control input 
$u_{\pw}(x, \mathcal{T}) = (u_{r, \pw}(x, \mathcal{T})^T, u_{s, \pw}(x, \mathcal{T})^T)^T$
is constructed as a feedback linearizing controller of the form \eqref{eq:FeedLinT} with $\mu$ structured as \eqref{eq:mu}, then $u_{s, \pw}(x, \mathcal{T}) = \bar{u}_{s, \pw}(\mathcal{X}, \mathcal{T})$.
}

\vp

\textit{Proof: }
Extending Theorem 1's proof, we need only show
\begin{equation*}
\small
    \begin{aligned}
        \begin{bmatrix}
        \dot{y}_{1, \pw}^d \\ 
        \ddot{y}_{2, \pw}^d
        \end{bmatrix} &=
        \begin{bmatrix}
        \dot{y}_{1, \pw}^d \\ 
        \ddot{y}_{2, \pw}^{r, d} \\ 
        \ddot{y}_{2, \pw}^{s, d}
        \end{bmatrix} &=
        \begin{bmatrix}
        \star \\ 
        \ddot{y}_{2, \pw}^{s, d}
        \end{bmatrix}.
    \end{aligned}
\end{equation*}
Including this 
component 
with the 3 components 
given in 
Theorem 1's proof yields $u_{\pw}(x, \mathcal{T}) = \begin{bsmallmatrix}  \star \\ \bar{u}_{s, \pw} \end{bsmallmatrix}$, showing $u_{s, \pw}(x, \mathcal{T}) = \bar{u}_{s, \pw}(\mathcal{X}, \mathcal{T})$.

\vp

\newsec{Prosthesis Result.}
Feedback linearization can be performed on the prosthesis subsystem with limited information of the human system and yield the \textit{same} control law as when a controller is constructed for the prosthesis with full knowledge of the human dynamics and states, yielding full system stability.
Although the domains are fully-actuated, zero dynamics exist due to the relative degree 1 output. Since we constructed the outputs to satisfy PHZD conditions, the zero dynamics are stable. Hence, the theorems presented in this paper still apply.

\section{RESULTS AND CONCLUSIONS} \label{sec:ResultsConclusion}
\newsec{Simulation Results.}
An amputee-prosthesis model was built according to the procedure in \ref{sec:AmpProsth} for a 65.8 kg 1.73 m female. Gaits were generated for this model according to \ref{ssec:GaitGen} and the hybrid system \ref{ssec:HybridSyst} was simulated for 100 steps
with $u_{\pt}(x)$ and $u_{\pw}(x, \mathcal{T})$ acting on the full system for $\Domain_{\pt}$ and $\Domain_{\pw}$, respectively. The gaits were simulated again for 100 steps with $u_{r, \pt}(x)$ and $\bar{u}_{s, \pt}(\mathcal{X})$ acting on the human and prosthesis joints, respectively, for $\Domain_{\pt}$, and $u_{r, \pw}(x, \mathcal{T})$ and $\bar{u}_{s, \pw}(\mathcal{X}, \mathcal{T})$ for $\Domain_{\pw}$. Fig. 4 depicts the prosthesis control inputs for each simulation, showing they match identically, demonstrating Theorem 1 and 2. The control laws enforced the trajectories found from the optimization and yielded stable amputee-prosthesis walking in simulation, as can be seen in \url{https://youtu.be/eFZ2_s6FGUI}.

The prosthesis subsystem was simulated for the continuous domains with values of $\bar{\theta}_B$, $\dot{\bar{\theta}}_B$, and $F_f$ obtained from the full system simulation. 
Additionally, during $\Domain_{\pw}$, $\tau_{\pw}$, $\dot{\tau}_{\pw}$, and $\ddot{\tau}_{\pw}$ from the full system were fed to the prosthesis simulation. 
These subsystem states from the prosthesis simulation and full system simulation are shown in Fig. 5, again showing they match, demonstrating Theorem 2. These phase portraits also show the prosthesis followed stable periodic orbits.
A second model was modified with human parameters increased by 24.9 kg and simulated using the same trajectories. 
Since the prosthesis only relies on the force measurement and global position and velocities, it still achieved perfect tracking of the outputs, shown in Fig. 6.

\newsec{Conclusions.}
This paper presented a novel framework for controlling separable systems, resulting in a controller for the subsystem, equivalent to a controller with knowledge of the full states and dynamics.
This formulation enables users to develop model dependent controllers for subsystems with limited information of the full system, with the same full system stability guarantees. Further, we outlined how to isolate a subsystem from an open-chain manipulator. This decomposition allows control of robotic modules without knowledge of its full system and provides a modeling method for robots that interact with another dynamic system. We demonstrated these methods on a prosthesis.
These results could, therefore, yield new nonlinear control approaches to prostheses that give formal guarantees on stability and safety while being applicable to real-world devices.

\begin{figure} \label{fig:SubsystemControlInputs}
\centering
\includegraphics[width=0.38\textwidth]{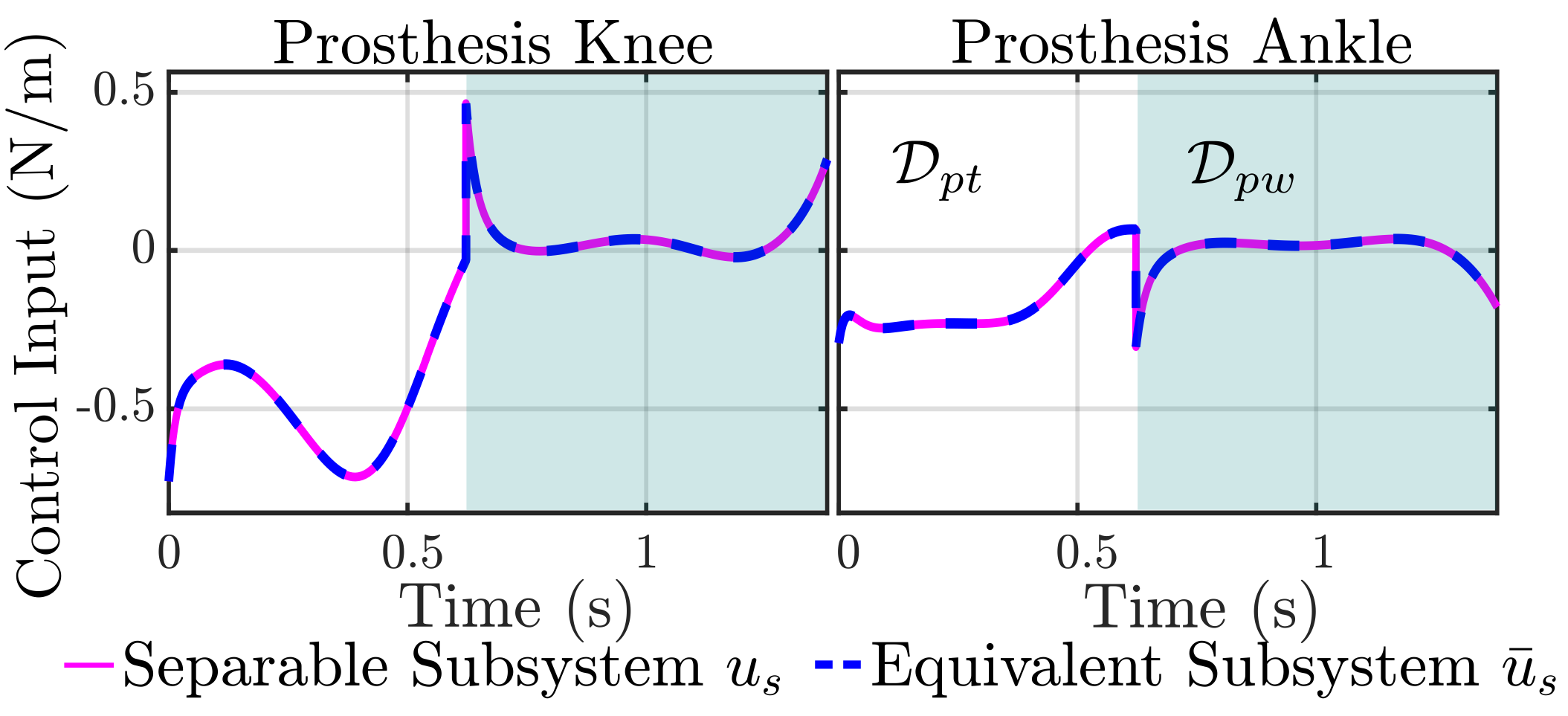}
\vspace{-0.4cm}
{\caption{Control inputs of prosthesis knee (left) and prosthesis ankle (right) for subsystem control law $u_s$ and $\bar{u}_s$ over stance and nonstance domains.}}
\vspace{-0.4cm}
\end{figure}

\begin{figure} \label{fig:SubsystemPhasePortraits}
\centering
\includegraphics[width=0.42\textwidth]{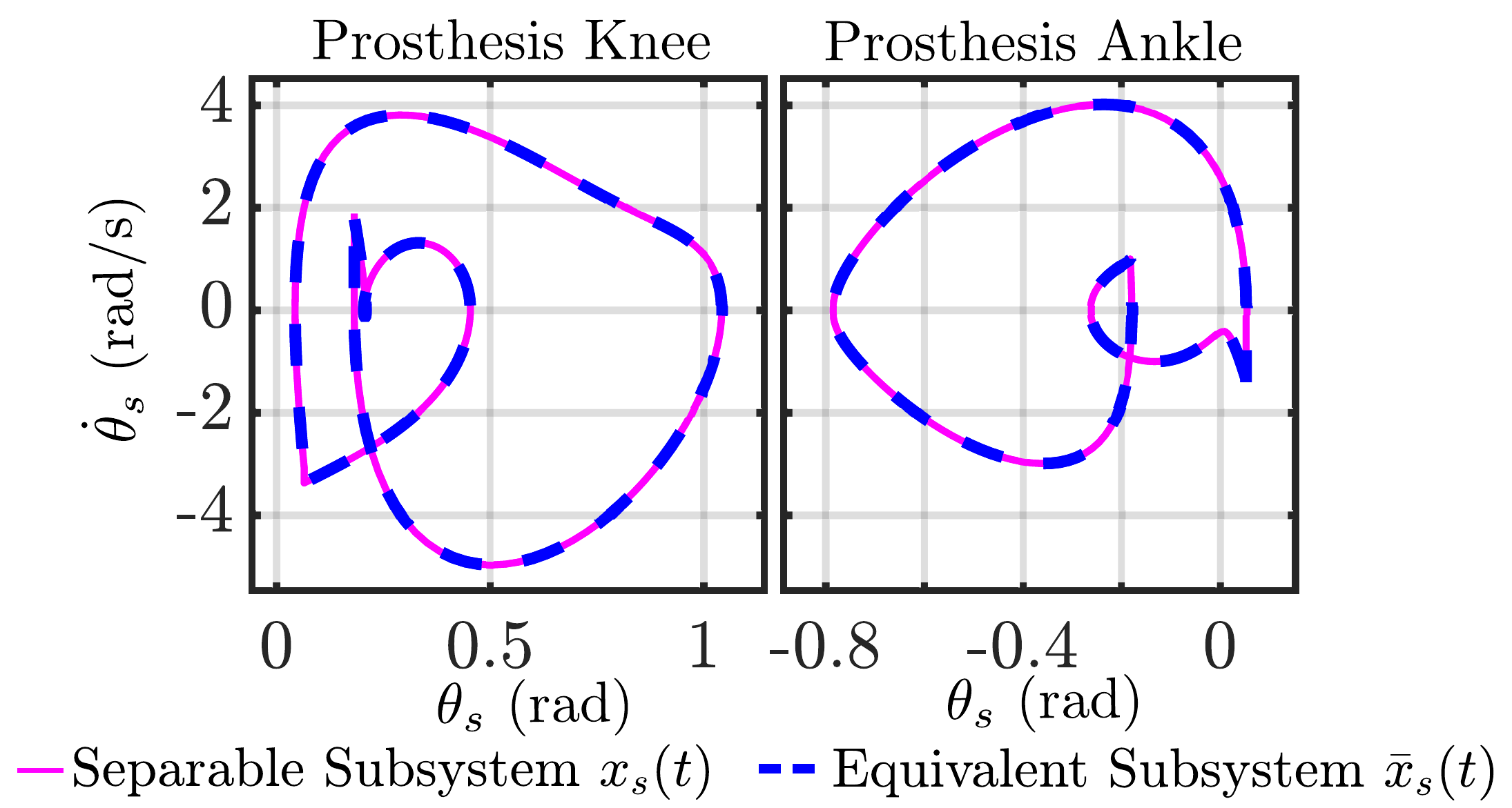}
\vspace{-0.4cm}
{\caption{Phase portraits of prosthesis knee (left) and prosthesis ankle (right) for stance and nonstance domains.}}
\vspace{-0.7cm}
\end{figure}

\begin{figure} [t] \label{fig:Models_OutputsEdit}
\centering
\includegraphics[width=0.41\textwidth]{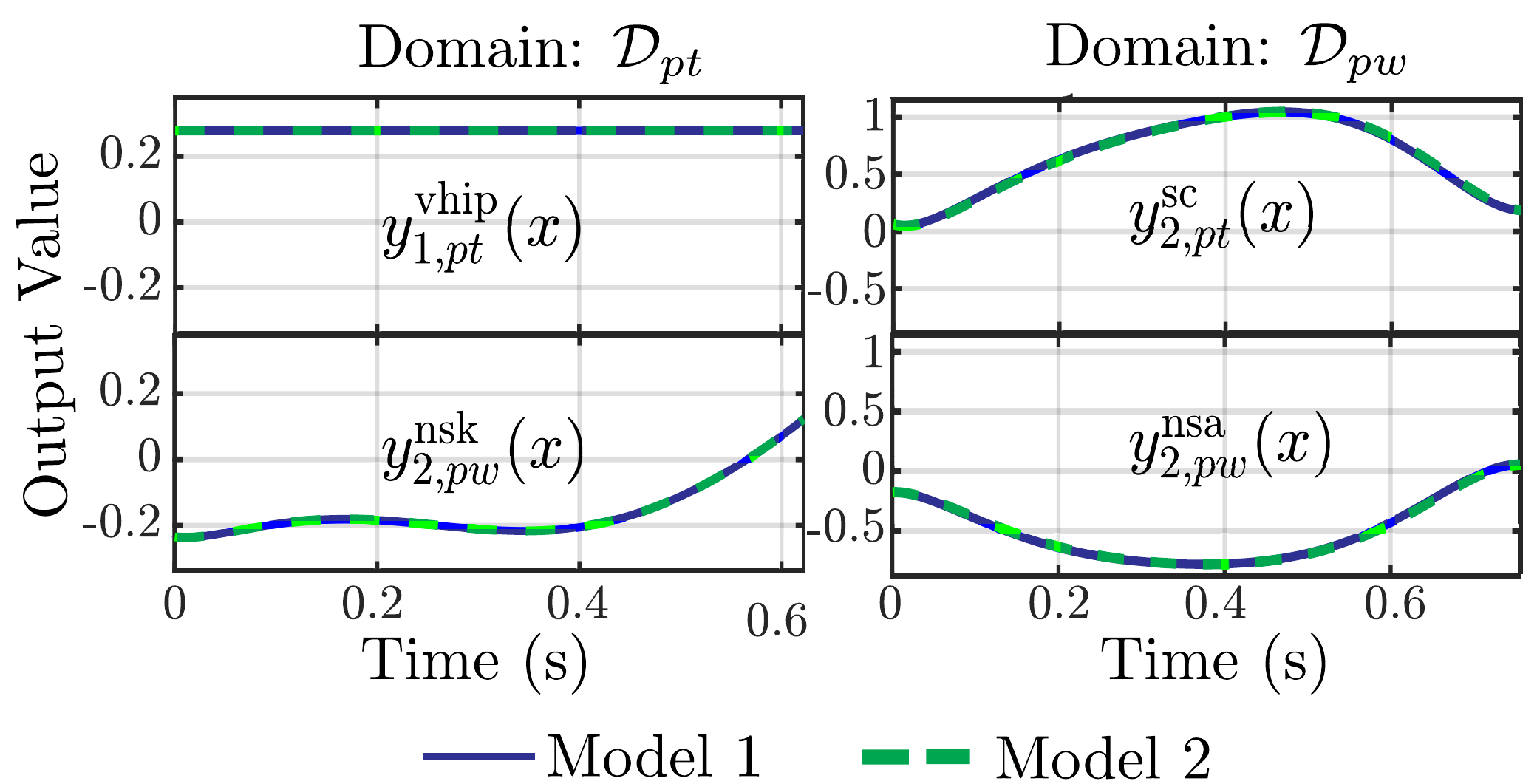}
\vspace{-0.4cm}
{\caption{Output functions of prosthesis knee and ankle for $\Domain_{\pt}$ (right) and $\Domain_{\pw}$ (left) for 65.8 kg Model 1 and 90.7 kg Model 2.}}
\vspace{-0.8cm}
\end{figure}

\vspace{-0.3cm}

\bibliographystyle{ieeetr}
\bibliography{sample}

\end{document}